\pdfoutput=1
\documentclass[11pt]{article}

\usepackage{acl}

\usepackage{times}
\usepackage{latexsym}

\usepackage[T1]{fontenc}
\usepackage[utf8]{inputenc}

\usepackage{microtype}

\usepackage{inconsolata}

\usepackage{booktabs}
\usepackage{tabularx}
\usepackage{stackengine}
\usepackage{tikz}
\usepackage{makecell}
\usepackage{xcolor,colortbl}

\newcommand{\xlmr}{\mbox{XLM-R}}
\newcommand{\xmod}{\mbox{X-MOD}}
\newcommand{\swissbert}{\mbox{SwissBERT}}
\newcommand{\bertscore}{\mbox{BERTScore}}
\newcommand{\canine}{\mbox{\textsc{Canine}}}
\newcommand{\canines}{\mbox{\textsc{Canine-S}}}

\title{Modular Adaptation of Multilingual Encoders to \\Written Swiss German Dialect}

\author{Jannis Vamvas ~~~~~Noëmi Aepli ~~~~~Rico Sennrich
\vspace{0.1cm}\\
 Department of Computational Linguistics, University of Zurich\\
 \texttt{\{vamvas,naepli,sennrich\}@cl.uzh.ch}
}

\begin{document}
\maketitle
\begin{abstract}
Creating neural text encoders for written Swiss German is challenging due to a dearth of training data combined with dialectal variation.
In this paper, we build on several existing multilingual encoders and adapt them to Swiss German using continued pre-training.
Evaluation on three diverse downstream tasks shows that simply adding a Swiss German adapter to a modular encoder achieves 97.5\% of fully monolithic adaptation performance.
We further find that for the task of retrieving Swiss German sentences given Standard German queries, adapting a character-level model is more effective than the other adaptation strategies.
We release our code and the models trained for our experiments.\footnote{\url{https://github.com/ZurichNLP/swiss-german-text-encoders}}
\end{abstract}

\section{Introduction}
When applying natural language processing~(NLP) techniques to languages with dialectal variation, two typical challenges are a lack of public training data as well as varying spelling conventions.
In the case of Swiss German, which is spoken by around~5~million people and is often used for informal written communication in Switzerland, these factors make it more challenging to train a BERT-like text encoder for written text.

In this paper, we adapt pre-trained multilingual encoders to Swiss German using continued pre-training on a modest amount of Swiss German training data.
We evaluate the approaches on part-of-speech~(POS) tagging with zero-shot cross-lingual transfer from Standard German~\cite{aepli-sennrich-2022-improving}, as well as dialect identification~\cite{zampieri-etal-2019-report} and cross-lingual sentence retrieval based on a parallel Standard German--Swiss German test set~\cite{aepli-etal-2023-benchmark}.

We find that depending on the multilingual encoder, continued pre-training leads to an average improvement of 10\%–45\% in average accuracy across the three downstream tasks.
We then focus on comparing monolithic adaptation, where all the parameters of the encoder are updated during continued pre-training, to modular adaptation with language-specific modular components (\textit{language adapters}; \citealp{pfeiffer-etal-2022-lifting}).
Even though modular adaptation only updates a fraction of the parameters, it is competitive to monolithic adaptation.
Given these findings, we propose to extend the \swissbert{} model~\cite{vamvas-etal-2023-swissbert}, which was trained on Standard German and other languages, with a Swiss German adapter~(Table~\ref{tab:figure-1}).

\begin{table}[t]
\footnotesize
\renewcommand{\arraystretch}{2}
\renewcommand\tabularxcolumn[1]{m{#1}}
\newcolumntype{Y}{>{\centering\arraybackslash}X}
\begin{tabularx}{\columnwidth}{l|Y|Y|}
 & \textbf{Monolithic} & \textbf{Modular} \\ \hline
\rotatebox[origin=c]{90}{\textbf{~~Subwords~~}} & \cellcolor{red!2}\xlmr{}~$\rightarrow$ \href{https://huggingface.co/ZurichNLP/swiss-german-xlm-roberta-base}{Swiss~German \xlmr{}} &  \cellcolor{red!2}\xmod{}/\swissbert{}~$\rightarrow$ \href{https://huggingface.co/ZurichNLP/swissbert}{Swiss German adapter} \\ \hline
\rotatebox[origin=c]{90}{\textbf{~~Characters~~}} &  \cellcolor{red!2}\canine{}~$\rightarrow$ \href{https://huggingface.co/ZurichNLP/swiss-german-canine}{Swiss~German \canine{}} &  \cellcolor{red!2}\xmod{}/\swissbert{}~$\rightarrow$ \href{https://huggingface.co/ZurichNLP/swiss-german-swissbert-char}{Swiss German character-level adapter} \\ \hline
\end{tabularx}
\caption{Overview of the encoder models we release.}
\label{tab:figure-1}
\end{table}

We further hypothesize that the architecture of \canine{}~\cite{clark-etal-2022-canine}, a tokenization-free model that operates on characters, might be better suited to the highly variable spelling of Swiss German.
Indeed, a \canine{} model adapted to Swiss German excels on the retrieval tasks, while POS tagging works better with subwords.

Finally, we aim to combine the best of both worlds by integrating character-level down- and upsampling modules into a subword-based model and training a \textit{character-level adapter} for Swiss German.
However, this jointly modular and tokenization-free strategy underperforms the individual approaches.
We hope that our findings can inform the development of modular approaches for other languages with dialectal variation.

\section{Adaptation Scenario}
Our goal is to train an encoder model for Swiss German~(language code~\texttt{gsw}) with limited training data.
Since Standard German~(language code \texttt{de}) is a closely related language, we focus on transfer learning from Standard German to Swiss German.
We rely on pre-trained multilingual models that have already been trained on Standard German, and adapt them to Swiss German using continued pre-training.

\paragraph{Swiss German adaptation data} For training on Swiss German, we use the SwissCrawl corpus~\cite{linder-etal-2020-automatic}, which contains 11M tokens of Swiss German text extracted from the web.
The text in SwissCrawl exhibits some normalizations that eventual input text will not have, e.g., isolation of individual sentences, normalization of punctuation and emoji removal.
To diversify the training data, we extend the pre-training dataset with a custom collection of 382k Swiss German tweets.
In total, we use 18M tokens for pre-training on Swiss German.
Both datasets were automatically mined and may contain some text in other languages.

\paragraph{Standard German data} To promote transfer from Standard German to Swiss German later on, we include an equal part of Standard German data in the continued pre-training data.
We use a sample of news articles retrieved from the Swissdox@LiRI database, comparable to the data the \swissbert{} model has been trained on~\cite{vamvas-etal-2023-swissbert}.

\section{Monolithic Approaches}
 We evaluate a subword-based model and a character-based model, with and without continued pre-training on Swiss German.
 We call these models monolithic (non-modular), because the entire model is updated during continued pre-training.

\subsection{\xlmr{}}
We train \xlmr{}~\cite{conneau-etal-2020-unsupervised} with masked language modeling~(MLM). \xlmr{} was pre-trained on 100 languages, which include Standard German but not Swiss German.

\subsection{CANINE} The \canine{} model~\cite{clark-etal-2022-canine} was pre-trained on 104 languages, again including Standard German but excluding Swiss German.
Unlike \xlmr{}, \canine{} directly encodes character sequences and does not require a tokenizer at inference time.
This is achieved by extending the standard transformer architecture with character down- and upsampling modules.

The \textit{downsampling module} combines a single-layer blockwise transformer with strided convolution, which reduces the sequence length by a factor of $r=4$, where $r$ is a hyperparameter.
As a consequence, the standard transformer does not see every character individually, but only sees downsampled positions.
The \textit{ upsampling module}, which is needed for token-level tasks, mirrors the downsampling procedure and restores the original sequence length.
We refer to \citet{clark-etal-2022-canine} for a detailed description of the architecture.

\citet{clark-etal-2022-canine} describe two alternative approaches for pre-training: \canines{}, which uses a tokenizer to determine masked tokens and is similar to standard MLM, and \canine{}-C, which is an autoregressive character loss.
In our experiments, we use \canines{} with the \swissbert{} subword tokenizer to perform continued pre-training.

\newcommand{\std}[1]{\footnotesize{$\pm$#1}}
\setstackgap{L}{5pt}

\begin{table*}
\setlength{\defaultaddspace}{2pt}
\centering
\begin{tabularx}{\textwidth}{@{}Xllccc@{}}
\toprule
& \textbf{POS} & \textbf{GDI} & \multicolumn{2}{c}{\textbf{Retrieval}} & \textbf{Macro-Avg.} \\
& & & \textsc{gsw-be} & \textsc{gsw-zh} & \\
\midrule
\xlmr{}: & & & & &  \\
– without continued pre-training & 52.6\std{1.8} & 47.2\std{15.1} & 60.6 & 75.7 & 56.0  \\
– with continued pre-training & \underline{86.9\std{0.3}} & 62.1\std{0.8} & 91.1 & 96.0 & \underline{80.9}  \\
\addlinespace
\addlinespace
\canine{}: & & & & &  \\
– without continued pre-training & 46.7\std{1.3} & 59.0\std{0.6} & 92.8 & 94.8 & 66.5  \\
– with continued pre-training & 60.9\std{1.4} & 60.8\std{0.4} & \underline{96.4} & \underline{96.9} & 72.8  \\
\addlinespace
\addlinespace
\swissbert{}: & & & & &  \\
– \textsc{de} adapter without continued pre-training & 64.8\std{2.0} & 61.3\std{0.5} & 66.1 & 82.2 & 66.7  \\
– subword-level \textsc{gsw} adapter & 83.2\std{0.3} & 62.0\std{0.4} & 82.9 & 92.4 & 77.6  \\
– character-level \textsc{gsw} adapter & 41.5\std{0.9} & 51.9\std{1.3} & 35.6 & 42.6 & 44.2  \\
\bottomrule
\end{tabularx}
\caption{Comparison of different models on three downstream tasks: part-of-speech (POS) tagging accuracy, German dialect identification (GDI) F1-score, and cross-lingual sentence retrieval accuracy.
For the supervised tasks, we report the average and standard deviation across 5 fine-tuning runs.
Underlined results indicate the best performance for a task.
}
\label{tab:main-results}
\end{table*}

\section{Modular Approaches}

\subsection{\swissbert{}}
We base our adapter experiments on \swissbert{}~\cite{vamvas-etal-2023-swissbert}, a variant of \xmod{}~\cite{pfeiffer-etal-2022-lifting} that includes language adapters for Standard German, French, Italian and Romansh.
Compared to the original \xmod{} model, which was trained with language adapters for 81 languages, \swissbert{} has a custom SentencePiece vocabulary and word embeddings optimized for Switzerland-related text, and we assume that this is beneficial for continued pre-training on Swiss German.

\subsection{Subword-level Adapter for \swissbert{}}

We add a Swiss German adapter to \swissbert{} and freeze the parameters of the model except for the adapter modules during continued pre-training.
%The left-hand side of Figure~\ref{fig:adapter-schema} schematically illustrates the modules updated during continued pre-training.
We initialize the Swiss German adapter with the weights of the Standard German adapter and pre-train it on the Swiss German part of our dataset.
During fine-tuning on downstream tasks, we freeze the adapters and update the remainder of the model.

For this approach, we only use the Swiss German part of our pre-training corpus for continued pre-training, and not Standard German, since the modular architecture is expected to allow for cross-lingual transfer without continued pre-training on the source language.
Table~\ref{tab:training-speed} provides an overview of the languages used for each approach.

\subsection{Character-level Adapter for \swissbert{}}
Previous work has found that learning a custom subword segmentation and embeddings that are adapted to the vocabulary of the target language can improve performance~\cite{wang-etal-2019-improving,pfeiffer-etal-2021-unks,vamvas-etal-2023-swissbert}.
However, this limits the degree of modularity, and we thus investigate a tokenization-free approach as an alternative.
In this experiment, we discard \swissbert{}'s subword embeddings when training the Swiss German adapter, and instead add the downsampling and upsampling modules of the \canine{} architecture.\footnote{We term this approach \textsc{Globi}~(\textbf{G}ranular \textbf{Lo}calization of \textbf{Bi}directional Encoders).}

Adding these modules results in exactly the same architecture as \canine{}, except that we opt for byte embeddings instead of character hash embeddings.
\canine{} uses a hash embedding method that can map any Unicode code point to a fixed-size embedding.
Since Standard German and Swiss German are mainly written in Latin script and there are limited training data, we forgo the hash embedding and learn UTF-8 byte embeddings instead.
%The vocabulary size is thus 261 (256 byte tokens + 5 special tokens).

Using the \canines{} objective, we first pre-train the character modules on Standard German pre-training data.
We then continue pre-training the adapters and the joint character modules on both languages, while freezing the rest of the model.
During fine-tuning, we freeze the adapters and train the remainder, analogous to the subword-level experiment.

\begin{table*}[]
\setlength{\defaultaddspace}{2pt}
\centering
\begin{tabularx}{\textwidth}{@{}Xllccc@{}}
\toprule
& \textbf{POS} & \textbf{GDI} & \multicolumn{2}{c}{\textbf{Retrieval}} & \textbf{Macro-Avg.} \\
& & & \textsc{gsw-be} & \textsc{gsw-zh} & \\
\midrule
\swissbert{} subword-level \textsc{gsw} adapter:  & & & & &  \\
– only updating the adapter weights & 83.2\std{0.3} & 62.0\std{0.4} & 82.9 & 92.4 & 77.6 \footnotesize{(97.5\%)} \\
– also updating the word embeddings & 83.9\std{0.1} & 62.1\std{0.3} & 86.0 & 93.7 & 78.6 \footnotesize{(98.7\%)} \\
– updating all the weights & 85.7\std{0.3} & 63.1\std{0.3} & 86.6 & 93.4 & 79.6 \footnotesize{\phantom{.}(100\%)} \\
\bottomrule
\end{tabularx}
\caption{
Effect of modularity on continued pre-training:
Only updating the adapter weights during continued pre-training achieves 97.5\% of the accuracy of a monolithic baseline where we update all the parameters of \swissbert{}.}
\label{tab:freezing-results}
\end{table*}

\section{Evaluation}

\subsection{Part-of-Speech Tagging (POS)}
Following \citet{aepli-sennrich-2022-improving}, we evaluate our models on POS tagging with zero-shot cross-lingual transfer from Standard German. To train the models, we use the German HDT Universal Dependencies Treebank \cite{borges-volker-etal-2019-hdt} and test on a dataset introduced by \citet{hollenstein-aepli-2014-compilation}. We report accuracy across the 54~STTS tags \cite{STTS}.\footnote{We mask the \texttt{APPRART} gold tag, which is not included in the training tag set, when calculating accuracy.}
We rely on the provided word segmentation and label the first token (subword/character/byte) of each word.

\subsection{German Dialect Identification (GDI)}
The GDI task~\cite{zampieri-etal-2019-report} is based on transcripts of the ArchiMob corpus of spoken Swiss German~\cite{samardzic-etal-2016-archimob}. This dataset contains four dialects, namely, Bern, Basel, Lucerne, and Zurich regions, constituting four distinct classes. We report the weighted F1-score.

\subsection{Sentence Retrieval}
For evaluating cross-lingual sentence retrieval, we use human translations of the English \texttt{newstest2019} source dataset~\cite{barrault-etal-2019-findings} into different languages.
Translations into Standard German are provided by NTREX-128~\cite{federmann-etal-2022-ntrex}; translations into Swiss German are provided by \citet{aepli-etal-2023-benchmark} for two regions, Bern~(\texttt{gsw-be}) and Zurich~(\texttt{gsw-zh}).

For both Swiss German test sets, we report the top-1 accuracy of retrieving the correct translation among all 1,997 translations, given the Standard German equivalent.
Note that 100\% accuracy is not attainable, since \texttt{newstest2019} has a small number of duplicate or near-duplicate sentences.
Following an evaluation approach used for \swissbert{}~\cite{vamvas-etal-2023-swissbert}, we perform unsupervised retrieval with the \bertscore{} metric~\cite{zhang-etal-2020-bertscore}.
We average the hidden states across all encoder layers. In the case of the \canine{}-style models, we use only the transformer layers that represent the downsampled positions.

\section{Experimental Setup}

\paragraph{Continued pre-training}
We combine Swiss German and Standard German training data with a 1:1 ratio.
The resulting bilingual dataset contains 37M tokens in total, and we set aside 5\% for validation~(Table~\ref{tab:training-validation-splits}).
We set the learning rate to 1e-4 and select the best checkpoint based on the validation loss out of 10 epochs; otherwise we use the default settings of Hugging Face transformer's \href{https://github.com/huggingface/transformers/blob/ffbcfc0166a5413176dc9401dbe5d3892c36fff6/examples/pytorch/language-modeling/run_mlm.py}{MLM example script}.
We train the models on a Nvidia V100 GPU with 32GB of memory and adjust the batch size dynamically to fit the available memory.
With the subword-based models, we set the sequence length to 512.
With the \canine{}-style models, we use the default downsampling rate of $r=4$ and a sequence length of $r \times 512 = 2048$ tokens~(characters or bytes).

\paragraph{Fine-tuning}
For the downstream tasks that involve fine-tuning~(POS and GDI), we fine-tune the model with a learning rate of 2e-5 and a batch size of 16.
We train for 10 epochs and select the best checkpoint based on the validation accuracy.
We report average and standard deviation across 5 fine-tuning runs with different random seeds.

\section{Results}
Table~\ref{tab:main-results} presents a comparison of the different models on the three downstream tasks.
Continued pre-training is highly beneficial for written Swiss German, confirming previous work~\cite{muller-etal-2021-unseen,aepli-sennrich-2022-improving,aepli-etal-2023-benchmark}.
This finding extends to the \canine{} model, for which language-adaptive pre-training has not been tested before, to our knowledge.

The adapted \canine{} shows state-of-the-art performance on the retrieval tasks.
A simple ChrF baseline~\cite{popovic-2015-chrf} achieves only 90.9\% and 93.0\% accuracy on the two retrieval tasks, and both the original and the adapted \canine{} clearly surpass this baseline.
However, the \canine{} model has low accuracy on POS tagging, reflecting previous findings for named entity recognition~\cite{clark-etal-2022-canine}.
Future work could explore alternative strategies for token-level classification tasks.

While the monolithic \xlmr{} model performs best overall, we consider adding a subword-based Swiss German adapter to \swissbert{} a competitive alternative, with the number of trainable parameters reduced by 95\%~(see Table~\ref{tab:model-parameters} for a comparison of the model sizes).
% XLM: 278,295,186
% Adapter: 14,825,814
Table~\ref{tab:freezing-results} confirms that restricting the continued pre-training to the adapter weights conserves most of the accuracy, compared to updating all the parameters of \swissbert{}.

Finally, a character-level adapter, where character up- and downsampling modules are added to the model specifically for Swiss German, performs better than random but clearly worse than the standard approaches.
This indicates that while the transformer layers of a subword-based model bear some similarity to the downsampled positions in the \canine{} architecture, continued pre-training cannot completely bridge the gap between the two architectures.
Future work could pre-train a modular character-level model from scratch to further improve adaptability to new languages and dialects, while taking into account more recent findings regarding the optimal design of character-level modules for text encoding~\cite{tay2022charformer,cao-2023-best}.

\section{Conclusion}
We compared strategies for adapting multilingual encoders to Swiss German.
We found that the monolithic approach of continued pre-training \xlmr{} is a strong baseline.
Adding a Swiss German adapter to \swissbert{}, a model with a modular architecture, is a viable alternative.
Finally, adapting \canine{} on Swiss German works well for cross-lingual retrieval.
The four Swiss German encoder models we trained for our experiments will be made available to the research community.

\section*{Limitations}
Differences between the pre-trained models make a fair comparison more difficult.
The encoder models we compare have originally been pre-trained with different data and hyperparameters~(but never on Swiss German).
They also differ in their number of parameters and vocabulary sizes, as detailed in Table~\ref{tab:model-parameters}.
Furthermore, we use a single, standard set of hyperparameters for pre-training and for evaluation, respectively.
Optimizing these hyperparameters for each model individually could lead to further improvements.

Finally, the evaluation results show that it is challenging to perform GDI classification purely based on written text, as previously discussed by~\citet{zampieri-etal-2017-findings}.
In interpreting the results, we focus mainly on the other two tasks, but still report results for GDI to provide a complete picture.

\section*{Acknowledgements}
This work was funded by the Swiss National Science Foundation (project nos.~213976 and 191934).
We thank Stefan Langer for helpful advice on collecting the Swiss German tweet dataset, and Chantal Amrhein for the provision of test data.
For this publication, use was made of media data made available via Swissdox@LiRI by the Linguistic Research Infrastructure of the University of Zurich (see \url{https://t.uzh.ch/1hI} for more information).

\bibliography{bibliography}

\appendix

\setcounter{table}{0}
\renewcommand{\thetable}{A\arabic{table}}

% https://tex.stackexchange.com/a/294990
\newcommand{\ExternalLink}{%
    \tikz[x=1.2ex, y=1.2ex, baseline=-0.05ex]{%
        \begin{scope}[x=1ex, y=1ex]
            \clip (-0.1,-0.1)
                --++ (-0, 1.2)
                --++ (0.6, 0)
                --++ (0, -0.6)
                --++ (0.6, 0)
                --++ (0, -1);
            \path[draw,
                line width = 0.5,
                rounded corners=0.5]
                (0,0) rectangle (1,1);
        \end{scope}
        \path[draw, line width = 0.5] (0.5, 0.5)
            -- (1, 1);
        \path[draw, line width = 0.5] (0.6, 1)
            -- (1, 1) -- (1, 0.6);
        }
    }

\clearpage
\onecolumn

\section{List of Encoder Models}
\label{sec:trained-models}

\begin{table*}[htb!]
\centering
\begin{tabularx}{\textwidth}{@{}Xrrrc@{}}
\toprule
\textbf{Model} & \textbf{Total parameters} & \textbf{Trained} & \textbf{Vocabulary size} & \textbf{URLs (original$\rightarrow$adapted)} \\
\midrule
\xlmr{} & 278M\phantom{\textsuperscript{\textdagger}} & 278M\phantom{\textsuperscript{\textdagger}} & 250,002 & \href{https://huggingface.co/xlm-roberta-base}{\ExternalLink} $\rightarrow$ \href{https://huggingface.co/ZurichNLP/swiss-german-xlm-roberta-base}{\ExternalLink} \\
\canine{} & 132M\textsuperscript{\textdagger} & 132M\phantom{\textsuperscript{\textdagger}} & - & \href{https://huggingface.co/google/canine-s}{\ExternalLink} $\rightarrow$ \href{https://huggingface.co/ZurichNLP/swiss-german-canine}{\ExternalLink} \\
\addlinespace
\swissbert{} & & \\
\mbox{– subword-level adaptation} & 139M\textsuperscript{\textdaggerdbl} & 8M\phantom{\textsuperscript{\textdaggerdbl}} & 50,262 & \href{https://huggingface.co/ZurichNLP/swissbert}{\ExternalLink} $\rightarrow$ \href{https://huggingface.co/ZurichNLP/swissbert}{\ExternalLink} \\
\mbox{– character-level adaptation} & 123M\textsuperscript{\textdaggerdbl} & 38M\textsuperscript{\textdaggerdbl} & 261 & \href{https://huggingface.co/ZurichNLP/swissbert}{\ExternalLink} $\rightarrow$ \href{https://huggingface.co/ZurichNLP/swiss-german-swissbert-char}{\ExternalLink} \\
\bottomrule
\end{tabularx}
\caption{The main encoders trained in this work.
\textsuperscript{\textdagger}~Figure does not include the \canines{} output embeddings, which can be discarded after pre-training.
\textsuperscript{\textdaggerdbl}~Figure includes two adapters (Swiss German and Standard German).
}
\label{tab:model-parameters}
\end{table*}

\vfill

\section{Ablation Study: Custom Subword Vocabulary}
\begin{table*}[htb!]
\setlength{\defaultaddspace}{2pt}
\centering
\begin{tabularx}{\textwidth}{@{}Xllccc@{}}
\toprule
& \textbf{POS} & \textbf{GDI} & \multicolumn{2}{c}{\textbf{Retrieval}} & \textbf{Macro-Avg.} \\
& & & \textsc{gsw-be} & \textsc{gsw-zh} & \\
\midrule
\xlmr{}: & & & & &  \\
– \xlmr{} vocabulary & 86.9\std{0.3} & 62.1\std{0.8} & 91.1 & 96.0 & 80.9  \\
– custom \textsc{gsw} vocabulary & 60.3\std{0.4} & 60.0\std{0.6} & 64.2 & 79.9 & 64.1  \\
\midrule
\swissbert{} subword-level \textsc{gsw} adapter\textsuperscript{\textdagger}: & & & & &  \\
– \swissbert{} vocabulary & 83.9\std{0.1} & 62.1\std{0.3} & 86.0 & 93.7 & 78.6  \\
– custom \textsc{gsw} vocabulary & 23.7\std{2.3} & 56.9\std{0.6} & 65.6 & 77.3 & 50.7  \\
\midrule
\canine{}: & & & & &  \\
– \canines{} with \swissbert{} vocabulary  & 60.9\std{1.4} & 60.8\std{0.4} & 96.4 & 96.9 & 72.8  \\
– \canines{} with custom \textsc{gsw} vocabulary & 57.8\std{1.2} & 62.1\std{0.6} & 95.6 & 96.3 & 71.9  \\
\midrule
\swissbert{} character-level \textsc{gsw} adapter: & & & & &  \\
– \canines{} with \swissbert{} vocabulary  & 41.5\std{0.9} & 51.9\std{1.3} & 35.6 & 42.6 & 44.2  \\
– \canines{} with custom \textsc{gsw} vocabulary & 40.6\std{1.2} & 11.0\std{1.9} & 28.7 & 38.4 & 28.4  \\
\bottomrule
\end{tabularx}
\caption{In an ablation experiment, we create a custom subword vocabulary for our continued pre-training dataset using SentencePiece~\cite{kudo-richardson-2018-sentencepiece}.
For the subword-based models, we train a new embedding matrix while initializing it with lexically overlapping embeddings from the original model.
Using the custom vocabulary for Swiss German decreases performance on all downstream tasks, probably due to the limited amount of training data.
For the character-based models, we use the \canines{} objective with the custom vocabulary.
Surprisingly, the custom vocabulary decreases performance, possibly because it is less similar to the subword vocabulary originally used by~\citet{clark-etal-2022-canine} to train \canines{}.
\textsuperscript{\textdagger}~In this experiment, we update the embedding weights of \swissbert{} to enable a fair comparison.
}
\label{tab:vocab-results}
\end{table*}

\smallskip

\begin{table*}[htb!]
\centering
\begin{tabularx}{0.65\textwidth}{@{}Xrr@{}}
\toprule
\textbf{Vocabulary} & \textbf{Vocabulary Size} & \textbf{Compression Ratio} \\
\midrule
\xlmr{} vocabulary & 250,002 & 3.36 \\
\swissbert{} vocabulary & 50,262 & 3.37 \\
\mbox{Custom \textsc{gsw} vocabulary} & 50,262 & 4.17 \\
\bottomrule
\end{tabularx}
\caption{Comparison of the SentencePiece vocabularies involved in the above ablation study. We report the compression ratio as the number of characters per subword token in a tokenized sample of our continued pre-training dataset.}
\label{tab:tokenizer-characteristics}
\end{table*}

\vfill
\clearpage

\section{Model Training Details}
\label{sec:training-details}

\begin{table*}[htb!]
\centering
\begin{tabularx}{\textwidth}{@{}Xrr@{}}
\toprule
\textbf{Approach} & \textbf{Languages trained} & \textbf{Training samples per second} \\
\midrule
\xlmr{} continued pre-training & \textsc{gsw} + \textsc{de} & 88.9 \\
\canine{} continued pre-training & \textsc{gsw} + \textsc{de} & 149.6 \\
\swissbert{} character-level adapter & \textsc{gsw} + \textsc{de} & 127.1 \\
\midrule
\swissbert{} subword-level adapter: & & \\
– only updating the adapter weights & \textsc{gsw} & 215.3 \\
– also updating the word embeddings & \textsc{gsw} & 202.4 \\
– updating all the weights & \textsc{gsw} & 225.9 \\
\bottomrule
\end{tabularx}
\caption{Empirical training speed in terms of training samples per second.
Note that training speed is only comparable for models trained on the same languages, since the \textsc{de} samples are longer than the \textsc{gsw} samples.}
\label{tab:training-speed}
\end{table*}

\section{Pre-training Datasets}
\label{sec:dataset-statistics}

\begin{table*}[htb!]
\centering
\begin{tabularx}{\textwidth}{@{}Xrrrrc@{}}
\toprule
\textbf{Dataset} & \textbf{Language} &  \textbf{Time Range} & \textbf{Examples} & \textbf{Tokens} & \textbf{URL} \\
\midrule
SwissCrawl~\cite{linder-etal-2020-automatic} & \textsc{gsw} & until 2019 & 563,037 & 10,961,075 & \href{https://icosys.ch/swisscrawl}{\ExternalLink} \\
Swiss German Tweets & \textsc{gsw} & 2007--2018 & 381,654 & 7,259,477 & - \\
Swissdox Sample & \textsc{de} & 2021 & 409,572 & 351,643,710 & \href{https://t.uzh.ch/1hI}{\ExternalLink} \\
\bottomrule
\end{tabularx}
\caption{Details of the datasets from which we source data for continued pre-training.}
\label{tab:pretraining-datasets}
\end{table*}

\begin{table*}[htb!]
\centering
\begin{tabularx}{0.8\textwidth}{@{}Xrr@{}}
\toprule
\textbf{Split} & \textbf{Examples} (news articles / tweets / sentences) & \textbf{Tokens} \\
\midrule
Training \textsc{gsw} & 897,477 & 17,308,288 \\
Training \textsc{de} & 20,140 & 17,459,689 \\
Validation \textsc{gsw} & 47,214 & 912,264 \\
Validation \textsc{de} & 1,082 & 905,476 \\
\bottomrule
\end{tabularx}
\caption{Training and validation splits used for continued pre-training.}
\label{tab:training-validation-splits}
\end{table*}

\section{Evaluation Datasets}
\label{sec:evaluation-dataset-statistics}

\begin{table*}[htb!]
\centering
\begin{tabularx}{\textwidth}{@{}XrrXc@{}}
\toprule
\textbf{Dataset} & \textbf{Examples} & \textbf{Tokens} & \textbf{Citation} & \textbf{URL} \\
\midrule
POS \textsc{de} (train) & 75,617 & 13,655,973 & \citet{borges-volker-etal-2019-hdt} & \href{https://github.com/UniversalDependencies/UD_German-HDT}{\ExternalLink} \\
POS \textsc{de} (validation) & 18,434 & 324,848 & \citet{borges-volker-etal-2019-hdt} & \href{https://github.com/UniversalDependencies/UD_German-HDT}{\ExternalLink} \\
POS \textsc{gsw} (test) & 7,320 & 113,565 & \citet{hollenstein-aepli-2014-compilation} & \href{https://noe-eva.github.io/publication/acl22/GSW_test_set.zip}{\ExternalLink} \\
\midrule
GDI (train) & 14,279 & 112,707 & \citet{zampieri-etal-2019-report} & - \\
GDI (validation) & 4,530 & 33,579 & \citet{zampieri-etal-2019-report} & - \\
GDI (test) & 4,743 & 42,699 & \citet{zampieri-etal-2019-report} & - \\
\midrule
Retrieval \textsc{de} & 1,997 & 50,833 & \citet{federmann-etal-2022-ntrex} & \href{https://github.com/MicrosoftTranslator/NTREX/}{\ExternalLink} \\
Retrieval \textsc{gsw-be} & 1,997 & 53,119 & \citet{aepli-etal-2023-benchmark} & \href{https://github.com/textshuttle/dialect_eval}{\ExternalLink} \\
Retrieval \textsc{gsw-zh} & 1,997 & 54,501 & \citet{aepli-etal-2023-benchmark} & \href{https://github.com/textshuttle/dialect_eval}{\ExternalLink} \\
\bottomrule
\end{tabularx}
\caption{Dataset statistics for the downstream tasks.}
\label{tab:evaluation-datasets}
\end{table*}

\end{document}